\pdfoutput=1
\documentclass[11pt]{article}

\usepackage[]{ACL2023}

\usepackage{times}
\usepackage{latexsym}
\usepackage{mathtools}
\usepackage{amsthm,amsmath,amssymb,amsfonts,exscale,latexsym,float,eucal}
\usepackage[T1]{fontenc}
\usepackage[utf8]{inputenc}

\usepackage{microtype}

\usepackage{inconsolata}

\usepackage{algorithm}
\usepackage{algorithmic}

\usepackage{graphicx}
\graphicspath{{./}}

\usepackage{tabularx}
\usepackage{setspace}
\usepackage{booktabs}
\usepackage{tablefootnote}

\title{Task-Optimized Adapters \\
for an End-to-End Task-Oriented Dialogue System
}

\author{Namo Bang$^*$ \\
  \\\And
  Jeehyun Lee$^*$ \\
  Department of Artificial Intelligence, Sogang University, Korea\\
  \texttt{\{namo950815, jhlee22, mwkoo\}@sogang.ac.kr} \\
 \\\And
  Myoung-Wan Koo \\
  }

\begin{document}
\maketitle
\def\thefootnote{*}\footnotetext{These authors contributed equally to this work.}\def\thefootnote{\arabic{footnote}}
\begin{abstract}
Task-Oriented Dialogue (TOD) systems are designed to carry out specific tasks by tracking dialogue states and generating appropriate responses to help users achieve defined goals. Recently, end-to-end dialogue models pre-trained based on large datasets have shown promising performance in the conversational system. However, they share the same parameters to train tasks of the dialogue system (NLU, DST, NLG), so debugging each task is challenging. Also, they require a lot of effort to fine-tune large parameters to create a task-oriented chatbot, making it difficult for non-experts to handle. Therefore, we intend to train relatively lightweight and fast models compared to PLM. In this paper, we propose an End-to-end TOD system with Task-Optimized Adapters which learn independently per task, adding only small number of parameters after fixed layers of pre-trained network. We also enhance the performance of the DST and NLG modules through reinforcement learning, overcoming the learning curve that has lacked at the adapter learning and enabling the natural and consistent response generation that is appropriate for the goal. Our method is a model-agnostic approach and does not require prompt-tuning as only input data without a prompt. As results of the experiment, our method shows competitive performance on the MultiWOZ benchmark compared to the existing end-to-end models. In particular, we attain state-of-the-art performance on the DST task of 2.2 dataset.\footnote{Our code is available at \url{https://github.com/sogang-isds/TOATOD.git}}
\end{abstract}

\begin{figure*}[th]
\centering
\includegraphics[width=1.0\textwidth]{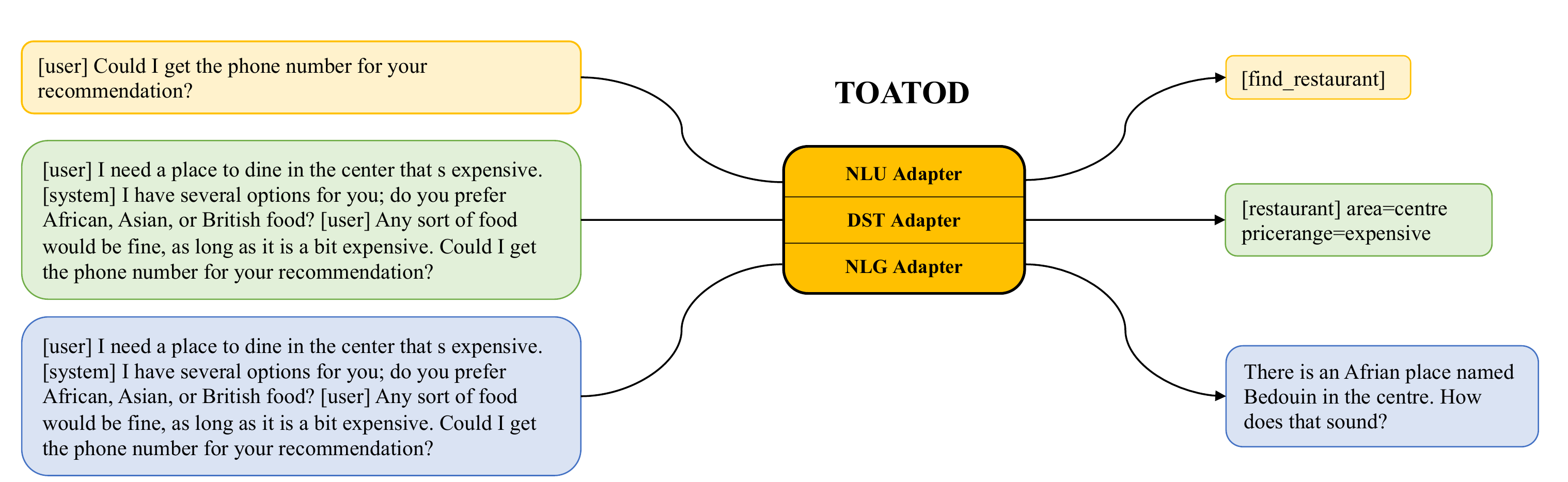} % Reduce the figure size so that it is slightly narrower than the column.
\caption{Overview of the Task-Optimized Adapters for an End-to-End Task-Oriented Dialogue System}
\label{fig:1}
\end{figure*}

\section{Introduction}

Task-oriented dialogue systems are trained to achieve specific goal to enhance efficiency and convenience in various fields such as customer service centers and healthcare information retrieval. Task-oriented dialogue systems are divided into key components: understanding the user's intent (NLU), tracking the current dialogue states (DST), and generating responses based on previous sessions (NLG). Pipeline-based systems separately train each component, so they have the advantage of optimizing each module and raising the performance of a given task. User feedback is, however, difficult to propagate to each module, and inputs to the component is dependent on the result of the previous module \citep{Chen2017}. Recently, dialogue systems have been trained in an end-to-end manner with transfer learning or pre-training networks with large dialogue corpora. However, building efficient end-to-end TOD systems requires a large amount of data and has some limitations due to parameter sharing. End-to-end models backpropagate to transfer the gradients of the output and end back to the entire neural network. They pose an issue of parameter efficiency as updating all parameters for every downstream scenario. Also, it is challenging to debug each task and take task-flow characteristics into account.

\setlength{\parskip}{0.2\baselineskip}
Therefore, we propose a simple structure of adding adapters to the core modules (NLU, DST, NLG) of the TOD system as shown in Figure \ref{fig:1}. By using the adapters, it is possible to optimize each task with only a small amount of training parameters, remaining the pre-trained model’s parameters fixed. Additionally, it is safe from the catastrophic forgetting problem \citep{forget}, which causes pre-trained models to lose important skills acquired during the pre-training process. The key is to apply a transfer learning strategy that yields compact and extensible downstream models in the dialogue system \citep{Houlsby2019}. This makes it easy for people to train large-scale end-to-end TOD models. Also, by applying REINFORCE \citep{rl}, we attempt to reduce the expected score gap caused by the small parameters of adapter compared to the full fine-tuning. Specifically, we use Joint Goal Accuracy, and weighted sum of BLEU and Success rate as rewards for training DST and NLG adapter. To best of our knowledge, this is first work that use Joint Goal Accuracy as a reward for E2E TOD system.

% \vspace{0.2\baselineskip}
To address the aforementioned problems, we propose a \textbf{T}ask-\textbf{O}ptimized \textbf{A}dapter for an end-to-end \textbf{T}ask-\textbf{O}riented \textbf{D}ialogue system (\textbf{TOATOD}) applying reinforcement learning to DST and NLG tasks. In summary, our key contributions are as follows:

\begin{itemize}
    \item We present a new architecture that can debug per task of the end-to-end model using the separated adapters.
    \item Without updating the original parameters of PLM, we train end-to-end TOD system efficiently with a few trainable parameters.
    \item It is a novel approach to design a reward function not just for NLG, but also for DST task with metric-aware reinforcement learning.
    \item The performance of the proposed approach outperforms on the DST task of MultiWOZ 2.2 and shows comparable results to full fine-tuning   on the NLU and NLG tasks.
\end{itemize}

\section{Background}
\noindent
\textbf{Pipeline-based Task-Oriented Dialogue System} Conventional task-oriented dialogue systems usually based on the pipeline method, consisted of language understanding (NLU), dialogue state tracking (DST), policy learning (POL), and language generation (NLG). This kind of modularization allows for each component to be optimized independently, making it easier to update and understand how the model is working. Pipeline-based systems,  however, have several limitations. Each of the modules train sequentially, so it is hard to align modules to the common optimization targets \citep{Liu2018}. This makes the system more complex and harder to backpropagate cumulated errors. The performance of the previous components affects the next modules, so if upper modules perform poorly, errors that occurred earlier may propagate and be amplified in downstream components \citep{Liu2018}.

% \vspace{0.2\baselineskip}
\noindent
\textbf{End-to-end Task-Oriented Dialogue System} End-to-end task-oriented dialogue systems, on the other hand, are easier to optimize and train to directly map the input to output in a single model. They can leverage large amounts of data for robust learning and the entire system can be optimized under end-to-end settings, which leads to better performance. A general approach for building end-to-end systems is to fine-tune pre-trained language models \citep{Budzianowski2019}. This approach utilizes the strength of pre-trained networks, which help the models to leverage the pre-trained knowledge while also adapting to task-specific data. For example, SimpleTOD \citep{simpletod} solved task-oriented dialogue as causal language modeling task using several versions of GPT \citep{Radford2018ImprovingLU, gpt}.

% \vspace{0.2\baselineskip}
\noindent
\textbf{Pre-training of Dialogue Language Model} Recently, methods with pre-training dialogue language model \citep{Wu2020, dialogpt, soloist, pptod, He2022}, instead of fine-tuning pre-trained networks have outperformed the previous baselines on the benchmark. For instance, SPACE-3 \citep{He2022} captures the contextualized knowledge from large-scale dialogue corpora by pre-training the unified language model. However, there are still some issues with these methods. A large amount of parameters is required for training backbone models like BERT, T5 \citep{t5}, GPT, and UniLM \citep{unilm}. As shown in Table \ref{table:Table 1}, 
$\text{T5}_{base}$ and $\text{T5}_{small}$ require the trainable parameters over 220M and 60M respectively. And they disregard the task-flow features of task-oriented dialogue systems. 
Also, it’s still hard to debug per module because the model parameters are shared and jointly optimized. PPTOD \citep{pptod} integrated modules into a unified model with task-specific prompts and alleviated the error accumulation in plug-and-play way. Still, this method is not completely free from the interference among tasks due to fully shared parameters.

\begin{figure}[t]
\centering
\includegraphics[width=1\linewidth]{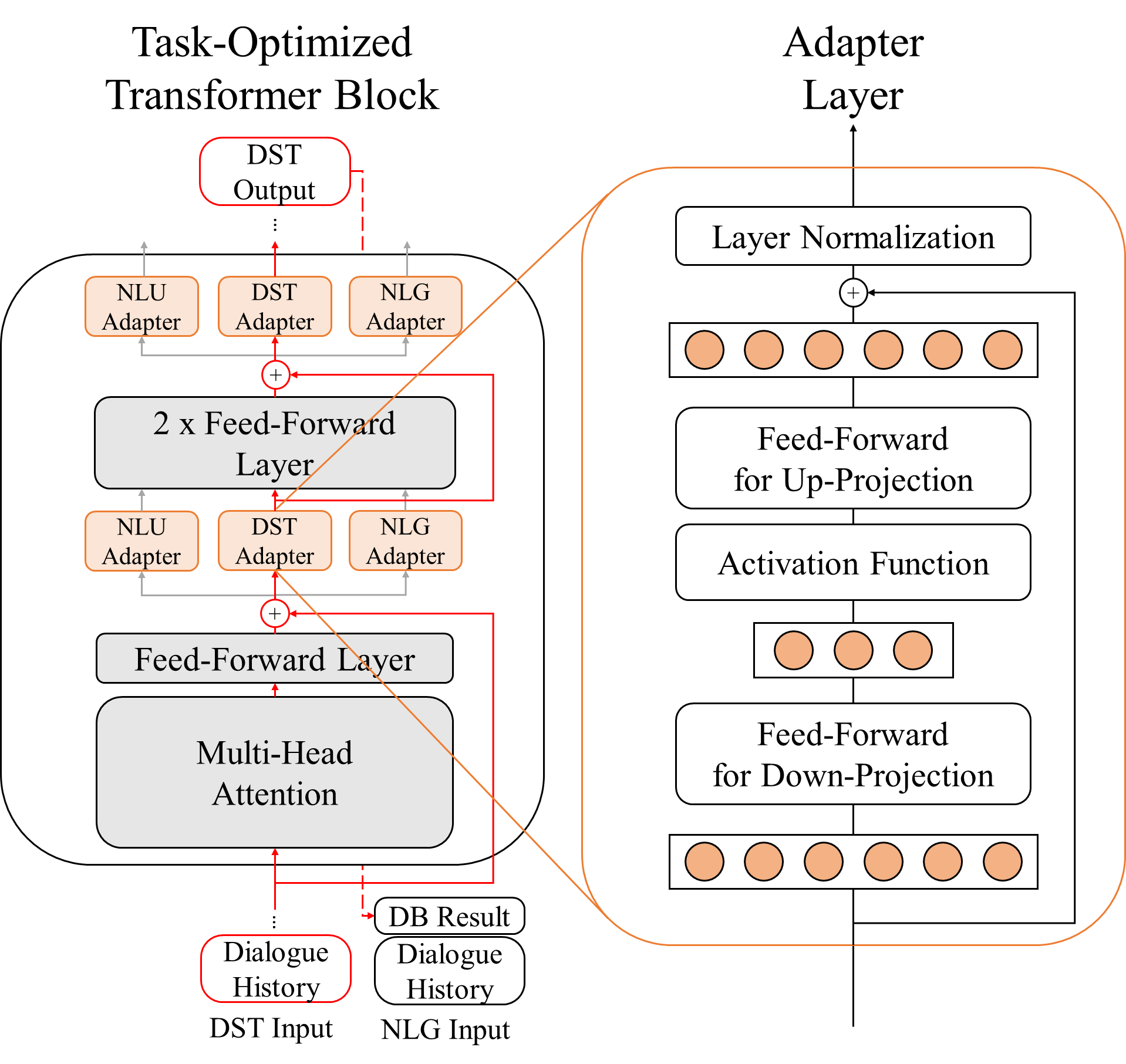} % Reduce the figure size so that it is slightly narrower than the column.
\caption{Architecture of the sub-modules in TOATOD. A red line indicates the forward path during DST inference. The DB result, based on the output of DST inference, is used for NLG inference sequentially. All of the adapter layers (NLU, DST, NLG) share the same architecture.}
\label{fig:2}
\end{figure}

% \vspace{0.2\baselineskip}
\noindent
\textbf{Adapter tuning for NLP} Since the pre-trained models for NLP tasks have become mainstream, they are mainly used for transfer learning downstream tasks. However, a parameter efficiency issue has been raised because updating all PLM parameters is expensive for every downstream scenario. To address the issue, the adapter module is proposed to transfer PLM like BERT with parameter-efficient tuning \citep{Houlsby2019} and shows comparable performance to full fine-tuning. The main idea behind the adapter is to train the network to the downstream task with task-specific parameters while maintaining the original pre-trained parameters. The module is composed of two feed-forward layers and a non-linear layer, which can be inserted into the transformer blocks of an end-to-end model. It projects $d$-dimensional input features into a smaller dimension $m$, and then projects back into the original dimension, so the total of parameters add per layer with biases is $2md + d + m$. The number of additional parameters per task can be restricted by setting $m < d$ \citep{adapter_survey}. In most of the previous research, the adapter was used for efficient learning. But in this approach, we adapt the adapter modules not just for efficient learning \citep{DBLP:conf/icml/Stickland019}, but also task-optimized learning as recent studies that have mainly focused on separating parameters \citep{adapterbot, adapter2, bapna-firat-2019-simple}.

% \vspace{0.2\baselineskip}
\noindent
\textbf{Reinforcement learning for Text Generation} Although token-based supervised learning is a widely adopted training method in text generation tasks, as highlighted by \citealp{Ranzato2015}, there are two major problems associated with this approach. The first problem is the exposure bias problem, where during training, the model is exposed to the ground-truth outputs, thus allowing it to learn to generate text that is similar to the training data. However, during evaluation, the model is not exposed to the ground-truth outputs and instead, generates texts based on the previous words generated by the model itself. This can lead to errors and a significant deviation in the generated text from the training data. The second problem is the token-level loss problem, where the training objective function is based solely on individual words' prediction, neglecting the predicted token sequence's overall coherence and fluency.

\setlength{\parskip}{0.2\baselineskip}
To address these problems, researchers have applied reinforcement learning to text generation tasks \citep{Ranzato2015, Li2016, Paulus2018, Lee2020, Wang2021, Ye22} using metrics such as BLEU \citep{Papineni2002} or ROUGE \citep{lin-2004-rouge} as rewards for sequence-level training. In our study, we also use the REINFORCE method and task-oriented dialogue metrics to continually train our model after supervised learning, not only to address these challenges but also to mitigate performance degradation caused by the use of an Adapter.

\begin{table}[h!]
\centering
\setlength{\tabcolsep}{2pt}
\renewcommand{\arraystretch}{1.7}
\scriptsize
\small
\begin{tabular}{lc llll}
\hline
               & \multicolumn{2}{l}{\textbf{Pre-trained}} & \multicolumn{2}{l}{\textbf{Trainable per task}} \\
\hline
\hline
\textbf{Model} & $\text{T5}_{base}$ & $\text{T5}_{small}$  & $\text{TOATOD}_{base}$ & $\text{TOATOD}_{small}$ \\
\textbf{\# of Prams} & 220M & 60M & 36M (14\%) & 7.9M (12\%) \\
\hline
\end{tabular}
\caption{This table shows the size of the pre-trained \text{T5} model (frozen shared parameters) and trainable Adapter per task in our models. We do not experiment with a large model, because the trainable parameter size of $\text{TOATOD}_{large}$ is bigger than $\text{T5}_{base}.$}
\label{table:Table 1}
\end{table}

\section{Methodology}

\subsection{Adapter for each task (NLU, DST, NLG)}

The Adapter module is designed to adapt the pre-trained network for each of the tasks (NLU, DST, NLG) in a task-oriented dialogue system. As illustrated in Figure \ref{fig:2}, the adapters are inserted after the feed-forward layer following the multi-head attention layer of the transformer blocks of the end-to-end model. It enables the model to learn task-specific representations while preserving the shared parameters learned during pre-training. As described in Table \ref{table:Table 1}, our model consists of large-size frozen shared parameters per task and small size of trainable parameters that account for about 14\% of the entire network. While original network’s parameters are frozen, the $j^{th}$ adapter of task $i \in $ \{NLU, DST, NLG\}, $A_{ij}$ computes as below: 

\begin{equation}
     \small
     {A}_{ij}=LN\ (W_{up} * \ ReLU\ \left(W_{down}*\ H_j\right)+\ H_j)
     \label{eq:1}
\end{equation}

The output of the $j^{th}$ feed-forward layer with residual connection in the transformer block is represented as $H_{j} \in \text{R}^{n\times d}$, where \text{n} is the input dimension, and \text{d} is the hidden dimension. As shown in Figure \ref{fig:2}, the overall architecture of the adapter module includes multiple feed-forward projections, referred to as down-projection and up-projection, followed by layer normalization (LN). The down-projection with $W_{down} \in \text{R}^{d\times h}$  projects the input $H_{j} $, which is passed by the ReLU activation function, and then the up-projection with $W_{up} \in \text{R}^{h \times d}$ projects the output back to the original dimension. The bottleneck dimension \text{h} is a hyperparameter to project the original input to a smaller dimension. And each adapter has a residual connection to avoid vanishing gradient \citep{DBLP:conf/nips/RebuffiBV17, DBLP:conf/cvpr/HeZRS16}.

\subsection{Metric-Aware Reinforcement Learning for DST \& NLG module}

The overall loss function for the metric-aware reinforcement learning is given by the equation (\ref{eq:2}):

\begin{equation}
   \small
    J\left(\theta \right)=\alpha \times J_{policy}\left(\theta \right)+\left(1-\alpha \right) \times CE\left({y,\hat{y}}\right)
    \label{eq:2}
\end{equation}

The equation describes how the network updates the parameters in order to maximize both the likelihood of the generated response (token loss) and the quality of the response (policy loss). The token loss can be described as a categorical cross-entropy loss. $\hat{y}$ denotes the predicted probability of ground truth and $y$ is the target probability. The token loss $CE(y,\hat{y})$, which measures how well a set of predicted token probabilities match the target tokens for a given context of dialogue when reinforcement learning. By applying REINFORCE method, the network can update weights towards the direction that allows the model for getting more rewards even when the reward function is non-differentiable.

\begin{equation}
   \small
    J_{policy}\left(\theta \right)={{-\ log} P({\hat{y}})\times Reward(y,\ \hat{y})\ } 
    \label{eq:3}
\end{equation}

The policy loss $J_{policy} (\theta)$  is introduced to measure of how well the model can generate a token sequence with high probabilities that result in high rewards. $\hat{y}$ denotes the predicted token sequence by model. The policy loss described in the equation (\ref{eq:3}) is calculated as the negative log probability of the token sequence that has the highest probability multiplied by the reward. 

The parameter $\alpha$ in the overall loss (\ref{eq:2}) is a hyperparameter, that is a scalar value between 0 and 1 to weigh the importance of these losses. In this way, the model is trained to predict the correct labels (categorical cross-entropy loss) and to make good decisions that result in high rewards (policy loss) at the same time. We define the reward functions of the DST and NLG modules as follows:

\begin{equation}
   \small
    {Reward}_{DST}=\ JGA(y,\ \hat{y})\ +\ 1
    \label{eq:4}
\end{equation}

The $Reward_{DST}$ is calculated as the sum of the Joint Goal Accuracy (JGA) and a constant value of 1. The JGA measures how well the model predicted the values for every slot in the dialogue turns. Using JGA as a reward, the model is encouraged to accurately track the state of the dialogue, which is crucial for generating appropriate responses and improving the performance of task-oriented dialogue systems.

{\small
    \begin{align}
    {Reward}_{NLG\ } = & (1-\beta )\times \mathbb{E}[BLEU (y_u,\ \hat{y_u})] \nonumber \\
    & +\ \beta \times Success(y,\hat{y}\ )\ +\ 1
    \label{eq:5}
    \end{align}}

In the equation (\ref{eq:5}), BLEU score and Success rate are used as rewards to guide the learning process of NLG module. $y_{u}$ denotes the ground truth token sequence of each utterance, and $\hat{y}_{u}$ denotes the predicted token sequence. To calculate the success rate, we apply the batch of a session-level, not an utterance-level. $y$ and $\hat{y}$ without $u$ mean the session level ground truth and prediction. The weighting factor \textbf{$\beta$} is adjusted to balance these two metrics. The hyperparameter $\beta$ may have to be carefully chosen because there is a trade-off relationship between these two metrics in the RL setting \citep{rl_setting}, where increasing one metric may come at the cost of decreasing the other. We experiment to choose \textbf{$\alpha$} and \textbf{$\beta$} on the Section 6.1.3 and 6.1.4.

\section{Experimental Setup}

\subsection{Datasets}

We experiment our method for dialogue state tracking (DST) and end-to-end response generation (NLG) tasks on the MultiWOZ 2.1 and 2.2 datasets, and the intent prediction (NLU) task with the Banking77, CLINC150, and HWU64 datasets.

% \vspace{0.3\baselineskip}
\noindent
\textbf{MultiWOZ - 2.1, 2.2} The MultiWOZ \citep{budzianowski-etal-2018-multiwoz} dataset has been widely used to evaluate the performance of TOD systems. It consists of 8438, 1000, and 1000 for training, dev, test sets with multi-turn dialogues, collected through a Wizard-of-Oz (WOZ) setup. The dialogues cover a wide range of domains and topics. MultiWOZ 2.2 \citep{multiwoz22} is the improved version of MultiWOZ 2.1 \citep{multiwoz21} that has corrected annotation errors, inconsistencies, and ontology issues, also added span annotations to standardize.  

% \vspace{0.3\baselineskip}
\noindent
{\textbf{Banking 77} \citep{banking77} This dataset is a collection of 77 real-life customer banking service queries. It consists of 13,083 utterances. Each query is labeled with a single intent, however, it is hard to differentiate because they correspond to very similar tasks.

% \vspace{0.3\baselineskip}
\noindent
\textbf{CLINC150} \citep{clinc150} This dataset is multi-domain dataset which contains 23,700 utterances that cover 150 intent classes over 10 domains. 

% \vspace{0.3\baselineskip}
\noindent
\textbf{HWU64} \citep{hwu64} This dataset consists of 25,716 examples. It maps user utterances to structured, but mode abstract. The data provides annotation with the 64 intents from 21 different domains.

\subsection{Baselines \& Settings}

\begin{table}
\centering
\renewcommand{\arraystretch}{2.3}
\resizebox{\linewidth}{!}{%
\begin{footnotesize}{\scriptsize}
\begin{tabular}{l|c|c|cl}
\cline{1-4}
\multicolumn{1}{c|}{ \Large{\textbf{Model}}} & \renewcommand{\arraystretch}{1}\begin{tabular}[c]{@{}c@{}} \large{\textbf{Backbone Model}} \\ \large{\textbf{(Trainable Prams)}}\end{tabular} & \renewcommand{\arraystretch}{1}\begin{tabular}[c]{@{}c@{}} \large{\textbf{MultiWOZ}} \\ \large{\textbf{2.1}}\end{tabular} & \renewcommand{\arraystretch}{1}\begin{tabular}[c]{@{}c@{}} \large{\textbf{MultiWOZ}} \\ \Large{\textbf{2.2}}\end{tabular}&  \\
\hline
\hline
\Large{TRADE}                      & -                                                                          & \Large{46.0}                              & \Large{45.4}                             &  \\ \cline{1-4}
\Large{DS-DST}                     & \Large${\text{BERT}}_{base}$ (110M)                                              & \Large{51.2}                              & \Large{51.7}                             &  \\
\Large{DST-Picklist}              & \Large${\text{BERT}}_{base}$ (110M)                                              & \Large53.3                              &                                  &  \\
\Large{TripPy}                     & \Large${\text{BERT}}_{base}$ (110M)                                              & \Large55.3                              &                                  &  \\
\renewcommand{\arraystretch}{1}\begin{tabular}[c]{@{}l@{}} \Large{{ConvBERT}} \\ \normalsize{{+DG+Multi}}\end{tabular}          & \Large${\text{BERT}}_{base}$ (110M)                                              & \Large58.7                              &                                  &  \\
\renewcommand{\arraystretch}{1}\begin{tabular}[c]{@{}l@{}} \Large{{Trippy}} \\ \normalsize{{+SaCLog}}\end{tabular}              & \Large$\Large{\text{BERT}}_{base}$ (110M)                                              & \Large60.61                             &                                  &  \\ \cline{1-4}
\Large{SimpleTOD}                  & \Large${\text{DistilGPT-2}}$ (82M)                                               & \Large56.45                             &                                  &  \\ \cline{1-4}
\Large{UniLM}                      & \Large${\text{UniLM}}$ (340M)                                                    & \Large54.25*                            & \Large54.25*                           &  \\
\Large{AG-DST}                     & \Large${\text{PLATO-2}}$ (310M)                                                  & \Large57.26                             & \Large57.26                            &  \\
\Large{SPACE-3}                    & \Large${\text{UniLM}}$ (340M)                                                    & \Large57.50                             & \Large57.50                            &  \\ \cline{1-4}
\Large{$\text{D3ST}_{base}$}                & \Large${\text{T5}}_{base}$ (220M)                                                & \Large54.2                              & \Large56.1                             &  \\
\Large{$\text{D3ST}_{large}$}               & \Large${\text{T5}}_{large}$ (770M)                                               & \Large54.5                              & \Large54.2                             &  \\
\Large{SDP-DST}                    & \Large${\text{T5}}_{base}$ (220M)                                                & \Large56.66                             & \Large57.60                            &  \\
\Large{$\text{PPTOD}_{base}$}                & \Large${\text{T5}}_{base}$ (220M)                                                & \Large57.10                             &                                  &  \\
\Large{$\text{PPTOD}_{large}$}               & \Large${\text{T5}}_{large}$ (770M)                                               & \Large57.45                             &                                  &  \\
\Large{$\text{D3ST}_{\text{XXL}}$}                 & \Large${\text{T5}}_\text{XXL}$ (11B)                                                  & \Large{\textbf{57.80}}                             & \Large58.7                             &  \\ \cline{1-4}
\Large{$\text{TOATOD}_{small}$}       & \Large{${\text{T5}}_{small}$ (7.9M)}                                               & \Large53.02                             &\Large{\textbf{61.92}}                            &  \\
\Large{TOATOD$_{base}$}        & \Large{${\text{T5}}_{base}$ (36M)}                                                 & \Large54.97                             & \Large{\textbf{63.79}}                            &  \\
\hline
\end{tabular}%
\end{footnotesize}
}
\caption{Joint Goal Accuracy for DST results. We use the best models after hyperparameter tuning and applying REINFORCE. Our models get the slightly higher score on the 2.2 dataset. Trainable Params denotes the number of trainable parameters of the network. The values with * are from SPACE-3 \citep{He2022}.}
\label{table:Table 2}
\label{tab:accents}

\end{table}
\begin{table*}[t]
\centering
\renewcommand{\arraystretch}{1.3}
\resizebox{\linewidth}{!}{%
\begin{tabular}{l|l|llll|llll}
\hline
\Large{Model}                                          & \renewcommand{\arraystretch}{1}\begin{tabular}{c}\Large{Backbone Model} \\ \large{(Trainable Params)}\end{tabular}                    & \multicolumn{4}{c}{\Large{MultiWOZ 2.1}}   & \multicolumn{4}{|c}{\Large{MultiWOZ 2.2}}    \\ \hline
\hline
                                               &                                                      & \Large{Inform} & \Large{Success} & \Large{BLEU}   & \Large{Combined} & \Large{Inform} & \Large{Success} & \Large{BLEU}  & \Large{Combined} \\ \hline
\Large{DoTS}                                           & \Large{${\text{BERT}}_{base}$ (110M)}   & \Large{86.65}  & \Large{74.18}   & \Large{15.90}  & \Large{96.32}    & -      & -       & -     & -        \\
\Large{PPTOD}                                          & \Large{${\text{T5}}_{base}$ (220M)}     & \Large{87.09}  & \Large{79.08}   & \Large{19.17}  & \Large{102.26}   & -      & -       & -     & -        \\
\Large{UBAR}                                           & \Large{${\text{GPT2}}$ (1.5B)}          & \Large{95.70}  & \Large{81.80}   & \Large{16.50} & \Large{104.94}   & \Large{83.4}   & \Large{70.3}    & \Large{17.6}  & \Large{94.4}     \\
\Large{MTTOD}                                          & \Large{${\text{T5}}_{base}$ (360.9M)*}  & \Large{90.99}  & \Large{82.08}   & \Large{19.68}  & \Large{106.22}   & \Large{85.9}   & \Large{76.5}    & \Large{19.0}  & \Large{100.2}    \\
\Large{RSTOD}                                          & \Large{${\text{T5}}_{small}$ (105.5M)*} & \Large{93.50*} & \Large{84.70*}  & \Large{19.24*} & \Large{108.34*}  & \Large{83.5}   & \Large{75.0}    & \Large{18.0}  & \Large{97.3}     \\
\Large{GALAXY}                                         & \Large{${\text{UniLM}}$ (340M)}         & \Large{95.30}  & \Large{86.20}   & \textbf{\Large{20.01}}  & \textbf{\Large{110.76}}   & \Large{85.4}   & \Large{75.7}    & \Large{19.64} & \Large{100.2}    \\
\Large{MinTL}                                          & \Large{${\text{BART}}_{large}$ (440M)}  & -      & -       & -      & -        & \Large{73.7}   & \Large{65.4}    & \Large{19.4}  & \Large{89.0}     \\
\Large{SOLOIST}                                        & \Large{${\text{GPT2}}$ (1.5B)}          & -      & -       & -      & -        & \Large{82.3}   & \Large{72.4}    & \Large{13.6}  & \Large{90.9}     \\
\Large{BORT}                                           & \Large{${\text{T5}}_{small}$  (60M)}     & -      & -       & -      & -        & \Large{85.5}   & \Large{77.4}    & \Large{17.9}  & \Large{99.4}     \\
\Large{Mars}                                           & \Large{${\text{T5}}_{small}$  (60M)}     & -      & -       & -      & -        & \Large{88.9}   & \Large{78.0}    & \textbf{\Large{19.9}}  & \textbf{\Large{103.4}}    \\ \hline
\Large{${\text{TOATOD}}_{small}$} & \Large{${\text{T5}}_{small}$ (7.9M)}    & \Large{92.10}  & \Large{80.40}   & \Large{18.29}  & \Large{104.54}   & \Large{85.80}  & \Large{74.00}   & \Large{18.00} & \Large{97.90}    \\
\Large{${\text{TOATOD}}_{base}$}  & \Large{${\text{T5}}_{base}$ (36M)}    & \textbf{\Large{97.00}}  & \textbf{\Large{87.40}}   & \Large{17.12}  & \Large{109.32}   & \textbf{\Large{90.00}}  & \textbf{\Large{79.80}}   & \Large{17.04} & \Large{101.94}   \\ \hline
\end{tabular}%
}
\caption{Inform, Success, BLEU, Combined Score for NLG results. All results of other models are cited from the official leaderboard. The values with * are from RSTOD. For the MultiWOZ 2.2 evaluation, we used our models trained on the MultiWOZ 2.1 after replacing the DST-optimized adapter with those trained on MultiWOZ 2.2.}
\label{table:Table 3}
\end{table*}

\noindent For the DST task, we compare our models with other strong baselines including TRADE \citep{trade}, DS-DST \citep{ds-dst}, DST-Picklist \citep{ds-dst}, TripPy \citep{Heck2020}, ConvBERT+DG+Multi \citep{convbert+multi}, TripPy+SaCLog \citep{trippy_saclog}, SimpleTOD \citep{simpletod}, AG-DST \citep{ag-dst}, UniLM, SPACE-3, D3ST \citep{d3st}, SDP-DST \citep{sdp-dst}, and PPTOD. On the NLG task, we choose models trained on PLM in an end-to-end setting such as DoTS \citep{dots}, PPTOD, UBAR \citep{ubar}, MTTOD \citep{mttod}, RSTOD \citep{rstod}, GALAXY \citep{galaxy}, MinTL \citep{mintl}, SOLOLIST, BORT \citep{bort} and Mars \citep{mars}. For the NLU task, we compare TOATOD with existing baselines of each dataset.

\setlength{\parskip}{0.2\baselineskip}
We use ${\text{T5}}_{small}$ and ${\text{T5}}_{base}$ as backbone models initialized with PPTOD \citep{pptod}'s pre-trained weights, which are trained on a large dialogue dataset. As described in the Appendix \ref{appendix:a}, the bottleneck dimension \text{h} of the adapter is 1/2 of the hidden dimension of the ${\text{T5}}$ model. We use the Adafactor \citep{Shazeer2018} optimizer with 15 epochs and set batch size as 16, learning rate of 1e-4 during supervised learning of DST and NLG tasks. We sweep a wide range of learning rates: ${\{}$1e-5, 1e-6, 1e-7${\}}$. For reinforcement learning, we train models 10 epochs for DST and 3 epochs for NLG. We do not train NLG-optimized adapters for the MultiWOZ 2.2, because there are not significant changes of response annotations from MultiWOZ 2.1 and intend to train robust models with noised dataset.

\setlength{\parskip}{0.2\baselineskip}
We follow the preprocessing method from UBAR to delexicalize slot values for each system responses. We evaluate our models using the older version of the standardized evaluation script for MultiWOZ 2.1, and the newly opened version for MultiWOZ 2.2, released by \citealp{mw-evaluation}. It has been adopted by the official MultiWoZ dataset github \footnote{\url{https://github.com/budzianowski/multiwoz}}. Other implementation details are described in Appendix \ref{appendix:c}.

\setlength{\parskip}{0.1\baselineskip}
\section{Experimental Results}
 
\subsection{Dialogue State Tracking}
% \vspace{0.5\baselineskip}
\noindent
We evaluate our models on the DST task with the MultiWOZ 2.1 \& 2.2 datasets. We compute Joint Goal Accuracy on the test set, which measures how many values are filled accurately compared to the ground truth states for all slots. Joint Goal Accuracy is considered as more difficult and important metric in most research \citep{jga1, jga2}, because once wrong prediction has been made, it cannot get the score at that turn.

\setlength{\parskip}{0.1\baselineskip}
\subsubsection{Evaluation Result}
\noindent We compare our best models, $\text{TOATOD}_{small}$ and $\text{TOATOD}_{base}$ to the models trained with a pre-trained network. Table \ref{table:Table 2} shows that our models are competitive to the end-to-end models on the current benchmark. In the 2.1 dataset, our models show a relatively good performance, despite the small number of trainable parameters. As shown in the Table \ref{table:Table 5}, the Joint Goal Accuracy of $\text{TOATOD}_{base}$ only with task-optimized adapter (SL) is 53.33, which is slightly lower than other models using ${\text{T5}}_{base}$ as backbone. So, we reduce the performance degradation applying metric-aware REINFORCE and report the final results. Trainable parameters of our models are less than 1/2 of the models with the smallest parameters. The result implies that the adapter module helps the network more adaptable to the DST task by only activating a few parts of the model. Because of relatively small parameters, our model is more robust to overfitting problem with confused labels. As mentioned in Section 4.1, MultiWOZ 2.2 is the cleaned version of 2.1 dataset, so the performance is better in the 2.2. Among the top results on the 2.2 dataset, our models obtain state-of-the-art performance. It demonstrates that \text{TOATOD} optimizes well on the given task remaining the prior knowledge learned from the pre-trained network.

\subsection{End-to-End Response Generation}

\noindent We test our methods with end-to-end response generation (NLG) task on the MultiWOZ 2.1 \& 2.2 as in DST evaluation. Four metrics are used to measure the quality of generated responses. We measure if the system provides the appropriate entity (Inform rate), answers all the requested information (Success rate), and responds fluently (BLEU score). And the Combined Score for end-to-end response generation is computed as 'BLEU+0.5 $\times$ (Inform+Success)'.  Under the end-to-end settings, the models have to predict proper dialogue states and then generate responses based on the states.

\subsubsection{Evaluation Result}

\noindent From Table \ref{table:Table 3}, our models achieve comparable results (1.x point different from SOTA model) in all datasets.  The adapter module helps each part of the model be fine-tuned independently, therefore we can optimize the DST and NLG task respectively. Our model performs well on NLG task based on the belief states from DST modules and base knowledge gained from the large-scale dialog dataset during pre-training. $\text{TOATOD}_{base}$ attains the best score on the Inform and Success rate. It shows that the reinforcement learning of our approach is effective for adjusting trade-off problem between BLEU score and others.

\subsection{Intent Classification}
\begin{table}[h]
\centering
\renewcommand{\arraystretch}{1.5}
\resizebox{\linewidth}{!}{%
\begin{tabular}{lllll}
\cline{1-4}
\multicolumn{1}{c|}{\Large{Model}}                            & \multicolumn{1}{l|}{\Large{Banking77}} & \multicolumn{1}{l|}{\Large{CLINC150}} & \Large{HWU64}                &  \\ \cline{1-4}
\hline
\hline
\multicolumn{1}{l|}{\Large{BERT-FIXED}}                       & \multicolumn{1}{l|}{\Large{87.19'}}     & \multicolumn{1}{l|}{\Large{91.79'}}    & \Large{85.77'}                &  \\
\multicolumn{1}{l|}{\Large{BERT-DG}}                          & \multicolumn{1}{l|}{\Large{91.75*}}     & \multicolumn{1}{l|}{\Large{95.98*}}    & \Large{90.89*}                &  \\
\multicolumn{1}{l|}{\Large{cist-dial (mslm)}}                 & \multicolumn{1}{l|}{\Large{91.17*}}     & \multicolumn{1}{l|}{\Large{95.80*}}    & \Large{91.36*}                &  \\
\multicolumn{1}{l|}{\Large{USE}}                              & \multicolumn{1}{l|}{\Large{92.81'}}     & \multicolumn{1}{l|}{\Large{95.06'}}    & \Large{91.25'}                &  \\
\multicolumn{1}{l|}{\Large{USE+CONVERT}}                         & \multicolumn{1}{l|}{\Large{93.36'}}     & \multicolumn{1}{l|}{\Large{97.16'}}    & \Large{92.62'}                &  \\
\multicolumn{1}{l|}{\Large{ConvBERT+Pre+Multi}}               & \multicolumn{1}{l|}{\Large{93.44*}}     & \multicolumn{1}{l|}{\Large{92.38*}}    & \Large{\textbf{97.11*}}                &  \\
\multicolumn{1}{l|}{\Large{SPACE 2.0}}                        & \multicolumn{1}{l|}{\Large{\textbf{94.77*}}}     & \multicolumn{1}{l|}{\Large{97.80*}}    & \Large{94.33*}                &  \\ \cline{1-4}
\hline
\multicolumn{1}{l|}{\Large{${\text{TOATOD}}_{small}$}} & \multicolumn{1}{l|}{\Large{92.40}}     & \multicolumn{1}{l|}{\Large{\textbf{98.45}}}    & \Large{90.42}                &  \\
\multicolumn{1}{l|}{\Large{${\text{TOATOD}}_{base}$}}  & \multicolumn{1}{l|}{\Large{92.17}}     & \multicolumn{1}{l|}{\Large{98.01}}    & \Large{90.79}                &  \\ \cline{1-4}
\end{tabular}%
}
\caption{Accuracy score (\%) on all three NLU task dataset with full training. The values with ' are from banking77 paper \citep{banking77}, and * are from the leaderboard for DialoGLUE paper \citep{dialoglue} and benchmark \footnote{\url{https://eval.ai/web/challenges/challenge-page/708/leaderboard/1943}}.}

\label{table:Table 4}
\end{table}
\def\thefootnote{3}\footnotetext{\url{https://eval.ai/web/challenges/challenge-page/708/leaderboard/1943}}\def\thefootnote{\arabic{footnote}}
\noindent We test our models on the NLU task with Banking77, CLINC150, and HWU64. Intent prediction is the task to identify the intent behind a given input. The task is normally framed as a classification problem, so we set the metric as turn accuracy.

\subsubsection{Evaluation Result}

\noindent From Table \ref{table:Table 4}, while our models do not achieve the highest score on NLU task, it is important to note that they perform well in relation to the size when compared to other models. This highlights the effectiveness of our task-optimized adapter approach in achieving a balance between model performance and efficiency.

\section{Further Analysis and Discussion}
\subsection{Reinforcement Learning}
\subsubsection{w/o Reinforcement Learning }

\begin{table}[h]
\centering
\setlength{\tabcolsep}{2pt}
\renewcommand{\arraystretch}{1.4}

\resizebox{\linewidth}{!}{%
\begin{tabular}{l|ll|llll}
\hline
Task       & \multicolumn{2}{c|}{DST}                                            & \multicolumn{4}{c}{NLG}                                                                                                                     \\ \hline
\hline
Metrics    & \small{\textbf{JGA}}   & \small{\textbf{Slot F1}} & \small{\textbf{Inform}} & \small{\textbf{Success}} & \small{\textbf{BLEU}}  & \small{\textbf{Combined}} \\ \hline
SL (2.2) & 62.92                           & 93.72                             & 85.30                            & 77.00                             & \textbf{18.44} & 99.59                             \\ \hline
RL (2.2) & \textbf{63.79} & \textbf{93.96}   & \textbf{90.00}  & \textbf{79.80}   & 17.04                           & \textbf{101.94}   \\ \hline
SL (2.1)  & 53.33                           & 91.68                             & 88.90                            & 81.40                             & \textbf{18.73} & 103.88                             \\ \hline
RL (2.1)  & \textbf{54.97} & \textbf{92.01}   & \textbf{97.00}  & \textbf{87.40}   & 17.12                           & \textbf{109.32}  \\ \hline
\end{tabular}%
}
\caption{Task performance of $\text{TOATOD}_{base}$ before and after applying REINFORCE. SL means supervised learning and RL means reinforcement learning. The test results of the $\text{TOATOD}_{small}$ is attached to Appendix \ref{appendix:b}.}
\label{table:Table 5}
\end{table}

\noindent Described on the Table \ref{table:Table 5}, after applying reinforcement learning, the performance of DST and NLG modules are enhanced. Models with reinforcement learning obtain best JGA on the DST task and get the highest Combined Score on the NLG task. The BLEU scores fall slightly but the Combined Score rise with Inform and Success, which means that incorporating reinforcement learning into the training process leads our model to complete tasks more efficiently. We conduct hyperparameter tuning only for the $\text{TOATOD}_{base}$, so there is some gap in the degree of performance improvement between $\text{TOATOD}_{small}$ and $\text{TOATOD}_{base}$.

\subsubsection{Hyperparameters of REINFORCE}
\begin{table*}[th]
\centering
\begin{tabular}{l|lllllllll}
\hline
{${\alpha}$}        & 1.0    & 0.7    & 0.9    & 1.0    & 1.0    & 0.3    & 0.5    & 0.7    & 1.0    \\ \hline
{${\beta}$}       & 0.4    & 0.5    & 0.5    & 0.5    & 0.6    & 0.7    & 0.7    & 0.7    & 0.7    \\ \hline
\hline
Inform   & \textbf{97.50}  & 97.40  & \textbf{97.50}  & \textbf{97.50}  & \textbf{97.50}  & 93.40  & 97.00  & 97.30  & \textbf{97.50}  \\
Success  & 85.70  & 86.40  & 85.90  & 85.70  & 85.80  & 85.40  & \textbf{87.40}  & 87.10  & 86.10  \\
BLEU     & 16.01  & 16.35  & 16.07  & 16.03  & 16.04  & \textbf{18.36}  & 17.12  & 16.52  & 16.10  \\
Combined & 107.67 & 108.25 & 107.77 & 107.63 & 107.69 & 107.76 & \textbf{109.32} & 108.72 & 107.90 \\ \hline
\end{tabular}%
\caption{Hyperparameter experiment with {${\alpha}$} and {${\beta}$} on the NLG task. We experiment $\text{TOATOD}_{base}$ on the MultiWOZ 2.1.}
\label{table:Table 6}
\end{table*}

\noindent The hyperparameter ${\alpha}$ is used for balancing the importance between cross-entropy loss and policy loss. We re-train DST and NLG modules by optimizing the mixed objective function to get higher rewards. The second hyperparameter ${\beta}$ is a scaling factor to control the trade-off between BLEU score and Success rate for NLG module. The BLEU score is calculated based on the number of matching n-grams between the generated text and the reference text. A higher score indicates that the generated text is more similar to the label text. Success rate measures the proportion of dialogues, where the model successfully completes the task. A higher success rate means the model is better at achieving the user’s goal of the dialogue. In some cases, however, the model may generate text that is close to the reference text (high BLEU score) but not relevant to the current dialogue (low Success rate). So, we aim to reduce the gap between two metrics via hyperparameter tuning.

\subsubsection{\textbf{$\alpha$} of DST-optimized adapter}
\begin{table}[h]
\centering
\renewcommand{\arraystretch}{1.1}
\begin{tabular}{l|l|l|l}
\hline
\multicolumn{1}{c|}{${\alpha}$} & 1.0   & 0.9   & 0.7   \\ \hline \hline
JGA                                   & \textbf{54.97} & 54.96 & 54.76 \\ \hline
Slot F1                               & 92.01 & \textbf{92.02} & 91.95 \\ \hline
\end{tabular}%
\caption{Hyperparameter experiment with {${\alpha}$} on the DST task. We test $\text{TOATOD}_{base}$ on the MultiWOZ 2.1.}
\label{table:Table 7}
\end{table}
\noindent In the Table \ref{table:Table 7}, the result of ${\alpha}$ experiment on DST task indicates that 1.0 yielded the best performance, suggesting the significance of the policy loss in the overall loss function. By maximizing policy loss, the model is encouraged to make decisions that result in high rewards, which improves the performance of the DST task.

\subsubsection{\textbf{$\alpha$} and \textbf{$\beta$} of NLG-optimized adapter}

\begin{figure}[h]
\centering
\includegraphics[width=1\linewidth]{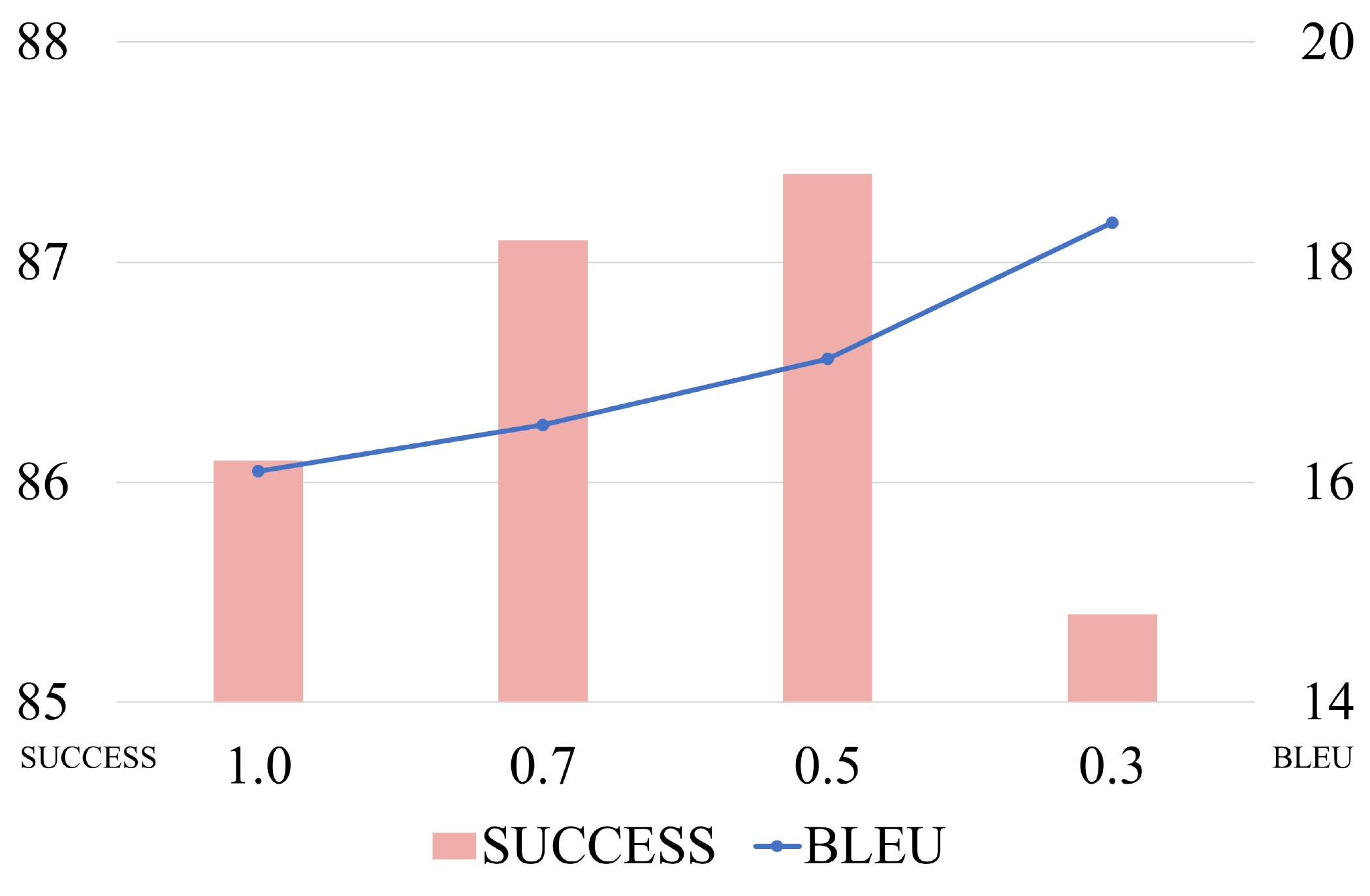} % Reduce the figure size so that it is slightly narrower than the column.
\caption{Effect of hyperparameter ${\alpha}$ when ${\beta}$ is fixed (${\beta}$=0.7). At the point of ${\alpha}$ is 0.5, trade-off issue occurs on MultiWOZ 2.1.}
\label{fig:3}
\end{figure}

\noindent We experiment with several combinations of ${\alpha}$ and ${\beta}$ within ${\{}$${\alpha}$: 0.3, 0.5, 0.7, 0.9, 1.0 / ${\beta}$: 0.4, 0.5, 0.6, 0.7${\}}$. Regardless of hyperparameters, performances improve after applying the reinforcement learning. The hyperparameters of ${\alpha}$=0.5 and ${\beta}$=0.7, result the best Combined Score, which is 5.44 point higher than the performance of the supervised learning. As shown in the Figure \ref{fig:3}, on the experiment after removing impact of CE loss (${\alpha}$=1), we found that the higher the ${\beta}$, success rate increases compared to BLEU score as we expected. 

As shown in Table \ref{table:Table 6}, when the ${\beta}$ is fixed, the bigger weight (smaller ${\alpha}$) on the CE loss ensures the higher BLEU score. On the other hand, the success rate started to decline from the certain point. Trade-off issue appears at this point. Therefore, ${\alpha}$ and ${\beta}$ need to be properly tuned in NLG task for good performance.

\section{Conclusion}

\noindent We propose \text{TOATOD}, task-optimized adapters for an end-to-end task dialogue system. By adapting task-optimized adapters, we utilize the end-to-end models without updating the pre-trained parameters and enabling debugging per task, which is different from previous research. In addition, we apply REINFORCE algorithm with metric-aware reward function directly not only on the NLG task but also DST task to prevent score degradation. As a result, we attain comparable performance to the previous SOTA models on every benchmark with very small number of trainable parameters. For the DST task of MultiWOZ2.2, our \text{TOATOD} model outperforms the current SOTA systems.

\section*{Limitations}
We train the task-optimized adapters based on the pre-trained weights of dialogue LM. Therefore, if applied to other dialogue tasks such as chit-chat and conversational QA system, the performance could be lower than that shown in our research. And we need future works to clarify the reason why the performance was better on the MultiWOZ 2.2 dataset, which is expected that our model does not overfit to the confused labels. Our model inferences in on the end-to-end manner, but trains like modular system for each task. End-to-end learning is currently under study. We could adapt the multi-task end-to-end learning to our method, which may lead to the better performance. Also, we could analyze the inner working of task-optimized adapters applying XAI technologies.

\section*{Ethics Statement}
We honor the ethical codes set out in the ACL code of Ethics. All of the datasets used in our study are from previous studies and do not have privacy issues.

\section*{Acknowledgements}
This work was supported by Institute of Information \& communications Technology Planning \& Evaluation (IITP) grant funded by the Korea government(MSIT) (No.2022-0-00621,Development of artificial intelli- gence technology that provides dialog-based multi-modal explainability)

\bibliography{anthology,custom}
\bibliographystyle{acl_natbib}

\appendix

\section*{Appendices}
\label{sec:appendix}

\section{Units of Adapters}
\label{appendix:a}

\begin{table}[h]
\centering
\renewcommand{\arraystretch}{1.4}
\small
\begin{tabular}{c|l|l|l}
\hline
Dim              & ${1}/{2}$                        & ${1}/{4}$                         & ${1}/{8}$          \\ \hline
JGA              & \multicolumn{1}{c|}{\textbf{51.59}} & \multicolumn{1}{c|}{51.11} & \multicolumn{1}{c}{50.43} \\
Slot f1          & 91.02                      & \textbf{91.07}                      & 90.40                     \\
Trainable Params & 7.9M                       & 7.2M                       & 6.6M                      \\ \hline
\end{tabular}%
\caption{Adapter units experiment results. We test different bottleneck dimension of adapter module on the DST task.}
\label{table:Table 8}
\end{table}
\noindent In this experiment, we evaluate the performance with several bottleneck dimensions, \text{h} = 256, 128, 64 with $\text{TOATOD}_{small}$, which is 1/2, 1/4, 1/8 size of $\text{T5}_{small}$'s embedding dimension of 512. The main focus of the experiment is to investigate the effect of the bottleneck dimension \text{h} of the adapter module on the performance of the model. To evaluate the performance of the model, we use the Joint Goal Accuracy for DST task. We keep the other hyperparameters constant across all the experiments, including learning rate of 1e-4 and evaluate the performance on the test set of MultiWOZ 2.1. We report the result on the Table \ref{table:Table 8}.

The result implies that there is a trade-off between the bottleneck dimension and the performance of the model. As the bottleneck dimension increases, the performance of the model also improves. The best performance is achieved with a bottleneck dimension of 256, where the JGA is 51.59. It is important to carefully choose the bottleneck dimension when using the adapter module in the task-oriented dialogue system. As described in the Table \ref{table:Table 1}, the trainable parameters of our model with bottleneck dimension of 256 is still significantly smaller than the PLM’s parameters, so we choose the size of 1/2.

\section{w/o Reinforcement Learning of $\text{TOATOD}_{small}$}
\label{appendix:b}

\begin{table}[h]
\centering
\setlength{\tabcolsep}{2pt}
\renewcommand{\arraystretch}{1.4}
\resizebox{\linewidth}{!}{%
\begin{tabular}{l|ll|llll}
\hline
Task       & \multicolumn{2}{c|}{DST}                                            & \multicolumn{4}{c}{NLG}                                                                                                                     \\ \hline
\hline
Metrics    & \small{\textbf{JGA}}   & \small{\textbf{Slot F1}} & \small{\textbf{Inform}} & \small{\textbf{Success}} & \small{\textbf{BLEU}}  & \small{\textbf{Combined}} \\ \hline
SL (2.2) & 61.29                           & 93.46                             & 78.80                            & 69.50                             & \textbf{18.46} & 92.61                             \\ \hline
RL (2.2) & \textbf{61.92} & \textbf{93.65}   & \textbf{85.80}  & \textbf{74.00}   & 18.00                           & \textbf{97.90}   \\ \hline
SL (2.1)  & 52.58                           & 91.31                             & 84.30                            & 74.40                             & \textbf{18.82} & 98.17                             \\ \hline
RL (2.1)  & \textbf{53.01} & \textbf{91.61}   & \textbf{92.10}  & \textbf{80.50}   & 18.28                           & \textbf{104.58}  \\ \hline
\end{tabular}%
}
\caption{Task performance of $\text{TOATOD}_{small}$ before and after applying REINFORCE. SL means supervised learning and RL means reinforcement learning.}
\label{table:Table 9}
\end{table}

\section{Implementation Details}
\label{appendix:c}

We used 8 A100 (80G) GPUs, but they were fully used during only reinforcement learning. During supervised learning, we used 4 GPUs. While reinforcement learning of the DST task-optimized adapters, we set learning rate as 1e-5, and batch size as 32 (utterance-level). On the contrary, while reinforcement training of the NLG task-optimized adapters, we set learning rate as 1e-6 and batch size as 4 (session-level). During the entire training process, we set random seed as 42. And for the NLU task, we used 1 RTX A5000 (24GB) GPU and trained models without reinforcement learning. We set batch size as 64 and single run with random seeds. When training $\text{T5}_{small}$ and $\text{T5}_{base}$ for the Banking 77 and CLINC150, we used learning rates of 0.001 and 0.15. And we used learning rates of 0.01 and 0.1 for the $\text{T5}_{small}$ and $\text{T5}_{base}$ with the HWU64.

\end{document}